# A comparative study of generative adversarial networks for image recognition algorithms based on deep learning and traditional methods

Yihao Zhong[1], Yijing Wei[2], Yingbin Liang[3], Xiqing Liu[4], Rongwei Ji[5], Yiru Cang[6]

[1]New York University, New York, USA

[2]Northwestern University, Evanston, USA

[3]Northeastern University, Seattle, USA

[4]Columbia University, New York, USA

[5]Southern Methodist University, Dallas, USA

[6]Northeastern University, Boston, USA

*Abstract:* In this paper, an image recognition algorithm based on the combination of deep learning and generative adversarial network (GAN) is studied, and compared with traditional image recognition methods. The purpose of this study is to evaluate the advantages and application prospects of deep learning technology, especially GAN, in the field of image recognition. Firstly, this paper reviews the basic principles and techniques of traditional image recognition methods, including the classical algorithms based on feature extraction such as SIFT, HOG and their combination with support vector machine (SVM), random forest, and other classifiers. Then, the working principle, network structure, and unique advantages of GAN in image generation and recognition are introduced. In order to verify the effectiveness of GAN in image recognition, a series of experiments are designed and carried out using multiple public image data sets for training and testing. The experimental results show that compared with traditional methods, GAN has excellent performance in processing complex images, recognition accuracy, and anti-noise ability. Specifically, Gans are better able to capture high-dimensional features and details of images, significantly improving recognition performance. In addition, Gans shows unique advantages in dealing with image noise, partial missing information, and generating high-quality images.

*Keywords: CNN, generative adversarial network, image recognition,*

## I. INTRODUCTION

In the digital era, image recognition has become a crucial research focus within computer vision. As digital image data grows and diversifies, the applications for image recognition expand, including image segmentation [1-3], object identification[4-6], medical image analysis[7-9], and disease diagnosis[10-13]. Progress in this area advances artificial intelligence and machine learning, offering numerous societal benefits. Despite notable successes, the field faces challenges such as handling complex images, noise interference, and varying image quality[14].

Earlier image recognition techniques primarily depended on manually crafted feature extraction algorithms and traditional machine learning models. Methods like Scale-Invariant Feature Transform (SIFT) and Histogram of Oriented Gradients (HOG) have been extensively used in these tasks. Although effective to an extent, these approaches struggle with the increasing volume and complexity of image data. The main drawbacks include limited feature representation, which hampers capturing high-level semantic details, and the need for extensive manual parameter tuning and feature engineering for varied and complex image tasks, leading to poor generalization.

Over the past few years, the swift evolution of deep learning technologies, notably the ascendancy of Convolutional Neural Networks (CNNs), has propelled significant advancements in image recognition. These networks facilitate an end-to-end learning approach that enables them to derive complex feature representations automatically from data, thereby bypassing the need for laborious, manual feature engineering inherent in conventional methodologies. A succession of milestone accomplishments in image recognition have been attained by CNN models, exemplified by AlexNet, VGGNet, GoogLeNet, ResNet, among others. These models have consistently dominated competitions such as the ImageNet Large-Scale Visual Recognition Challenge (ILSVRC), leading to a substantial enhancement in both the precision and speed of image recognition capabilities[15-18].

Despite the great success of deep learning technology in the field of image recognition, there are still some problems and challenges. For example, for some complex scenes or images with large noise interference, traditional deep learning models often perform poorly and are difficult to effectively identify. In addition, deep learning models often require a large amount of labeled data for training, the acquisition cost of labeled data is high, and there are problems with inaccurate and incomplete labeling. In order to solve these problems, researchers continue to explore and propose new methods and techniques.

Generative Adversarial Networks (GANs), a cutting-edge deep learning model, have garnered significant attention recently[19]. GANs consist of two neural networks: a generator and a discriminator. Through adversarial training, the generator learns to create realistic images, while the discriminator's role is to differentiate between real and generated images. This innovative approach has not only achieved remarkable success in image generation but also shown promise in enhancing image recognition. By incorporating GANs, the performance and robustness of recognition models can be significantly boosted, resulting in more accurate and comprehensive models. GANs contribute to improved model generalization and robustness by generating challenging image samples, which augment the training data. Additionally, GANs can learn the distribution characteristics of image data, aiding the recognition model in better capturing and understanding semantic information within images. This leads to enhanced accuracy and stability in recognition tasks.

This paper aims to deeply explore the image recognition algorithm based on the combination of deep learning and generative adversarial network, and make a comprehensive comparison and analysis with traditional image recognition methods. Through comparative experiments, the advantages and disadvantages of image recognition algorithms based on deep learning and generative adversarial networks in recognition accuracy, robustness, and data utilization efficiency are evaluated. At the same time, the potential application and future development trend of generative adversarial networks in the field of image recognition are also discussed. Through this research, it will provide important theoretical and practical references for further promoting the development of image recognition technology, improving the efficiency and accuracy of image recognition, and expanding the wide application of image recognition technology in practical applications.

## II. OVERVIEW OF TRADITIONAL IMAGE RECOGNITION METHODS

Convolutional Neural Networks (CNNs) have significantly advanced the field of image recognition, extending their application from simple classification tasks to a multitude of disciplines, including prediction algorithms [20-21], natural language processing [22-23], and speech recognition [24-25]. One of the earliest methodologies in image recognition, template matching, involves the correlation of a predefined template image with regions in a target image to ascertain the best match based on a predefined similarity metric. This technique was predominantly employed in applications characterized by limited variations in the appearance of the subject of interest.

Prior to the ascendancy of CNNs, feature-based recognition techniques were the mainstay. These methods focus on the identification, description, and subsequent matching of local features. Techniques such as the Scale-Invariant Feature Transform (SIFT) and Speeded Up Robust Features (SURF) have been particularly influential. These algorithms are designed to detect keypoints in an image and generate descriptors that are robust to scale, rotation, and partially robust to variations in illumination and perspective.

The Histogram of Oriented Gradients, introduced by Dalal and Triggs in 2005, represents another significant approach within traditional image recognition methodologies, primarily used for object detection. This technique involves dividing an image into numerous small, interconnected regions termed cells, and compiling a histogram of gradient directions or edge orientations within these cells. The aggregate of these histograms then forms a descriptor for the object.

Developed for efficient object detection, Haar Cascade classifiers employ Haar-like features to identify objects within an image or video sequence. Notably utilized for face detection, this algorithm processes image regions to calculate the relative differences in pixel intensities across adjacent regions. By cascading several stages of simple classifiers, this approach effectively discriminates between target objects and the background.

The Bag of Visual Words model, adapted from techniques in document classification, has been applied to both image retrieval and classification tasks. This model analogizes image features to words and images to documents, wherein features are extracted, encoded to construct a visual vocabulary, and subsequently used to generate a histogram depicting the distribution of these 'visual words' across the image. This methodology has proven particularly effective for applications in scene classification and texture recognition.

The transition from traditional image recognition methods to deep learning-based techniques, particularly CNNs, signifies a paradigm shift in the ability to autonomously learn hierarchical feature representations, greatly enhancing performance on complex and diverse image recognition tasks. This transition not only underscores the limitations inherent in traditional methods but also emphasizes the profound capabilities of contemporary deep learning models.

In our comparative analysis, these traditional methods are juxtaposed with modern deep learning approaches, such as Generative Adversarial Networks (GANs), to elucidate the substantial advancements within the field. This comparison serves not only to highlight the incremental improvements brought about by novel technologies but also to provide a comprehensive perspective on the current capabilities and potential future directions of image recognition technologies.

## III. GENERATIVE ADVERSARIAL NETWORK

### A. Generate adversarial network foundation The generative adversarial network consists of two neural network models

Generator $G$ and discriminator $D$. The learning process of GAN is the zero-sum game process of generator $G$ and discriminator $D$. The overall structure of GAN is shown in Figure 1.

Collaborative Filtering (CF) is a technique widely used in recommendation systems to predict user preferences for unknown items. The basic idea is that if two users have rated certain items similarly in the past, they are likely to have similar preferences for other items they have not yet encountered. CF algorithms are mainly divided into two categories: User-Based Collaborative Filtering (UBCF) and Item-Based Collaborative Filtering (IBCF) are shown in Figure 1.

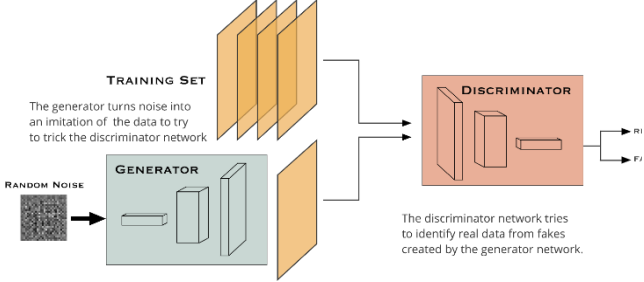

Figure 1 GAN network

Generated against the network operation process can be summarized into the following three steps: (1) Enter a random noise $z$ to generator $G$, generator $G$ through a false samples, the noise generated by $G(z)$, generator $G$'s main goal is to keep $D$ discriminant criterion device was born into a fake sample $G$ samples $(z)$ is false. (2) The real data $x$ of the sample and generator generated fake samples before $G(z)$ are input to the sentence don't implement $D$, output a probability value $D(x)$, if the output of $D(x)$ is 1, represents distinguishing device $D$ determining lost into the sample is a real sample, if the output $D(x)$ is zero. The discriminator $D$ evaluates an input sample to determine if it is a fake sample produced by the generator $G$. During training, if $D(x)$ is close to 1, it indicates a higher probability that the sample is real; if $D(x)$ is close to 0, it suggests the sample is likely a fake created by $G$. The primary objective of the discriminator is to distinguish between real data samples $x$ and fake samples $G(z)$ generated by $G$. As $G$ and $D$ undergo multiple rounds of alternate optimization, the fake samples $G(z)$ produced by $G$ become increasingly indistinguishable from real samples $x$. Eventually, the discriminator $D$ cannot reliably differentiate between real and generated samples, with its output $D(x)$ approaching 0.5 for both types. This indicates that the generator and discriminator have reached a dynamic equilibrium, known as Nash equilibrium.

The discriminator $D$ determines that the input sample is a fake sample produced by the synthesizer $G$. In the subsequent training, the closer $D(x)$ is to 1, the greater the possibility that the sample is from the real data; The closer $D(x)$ is to 0, the greater the possibility that the discriminant sample is from the fake sample generated by the generator $G$. The goal of the discriminator $D$ is to distinguish the sample $x$ from the real data and the fake sample $G(z)$ generated by the generator $G$. (3) After the generator $G$ and discriminator $D$ multiple alternate optimization after training, the generator $G$ generated fake samples $G(z)$ is more and more like a real sample $x$ $D$ discriminant device can samples out, whether from $x$ or real sample of real data from the generator $G$ generates a fake sample $G(z)$, The discriminator output $D(x)$ every time becomes 0.5, that is, the generator $G$ and the discriminator $D$ reach a dynamic equilibrium, also known as Nash equilibrium.

### B. Generate adversarial network training

The ultimate goal of Generative Adversarial Networks (GANs) is to input random noise $z$ (usually following a Gaussian distribution) into the generator $G$, which then outputs a generated sample $G(x;\theta)$ with a data distribution approximating that of real samples $P\text{data}(x)$, where $\theta$ represents the network parameters. The aim of training is to make the distribution $PG(x;\theta)$ closely match the distribution $P\text{data}(x)$. At the start of training, when random noise $z$ is input into the generator $G$, the generated sample $G(z)$ has a distribution that significantly differs from that of real samples since the generator is initially untrained. The discriminator $D$ receives both the fake samples $G(z)$ from the generator and the real samples $x$ from the dataset. Initially, the discriminator $D$ cannot effectively distinguish between fake and real samples. Therefore, training begins with the discriminator. Each training process can be summarized into two parts:

(1)The function of the $D$ discriminator is to distinguish between the samples received by $D(x)=1$ and $D(G(z))=0$, and the samples received by the $G$ generator (negative). In the training process of deep learning, the structure of cross-entropy, i.e., $log(x)$, is generally adopted for the objective function of this binary classification problem. So goals into making $log(D(x))$ to maximize, while making $log(D(G(z)))$ to minimize the $log(1-D(G(z)))$. Therefore, the maximum objective function of the discriminator $D$ in the training process is:

$$\max_D(D,G) = E_{x\sim P_{data(x)}}[logD(x)] \\ + E_{z\sim P_z(z)}[log(1-D(G(z)))] \quad (1)$$

Where, $Ex\sim Pdata(x)$ is real samples from real data set; $Ez\sim Pz(z)$ is fake samples generated from the generator $G$. $D(G(z))$ is the score of the discriminator $D$ on the fake sample $G(z)$ generated by the generator $G$ based on input noise $z$.

(2)Generator $G$ function is hoping to make $D$ discriminant criterion device is the key to the receiving samples is real samples or generated by their false samples, so the generator $G$ training goal is to make $D(G(z))$ the most exaggerated, according to the requirements of the cross entropy principle is $log(D(G(z)))$ to maximize, That is, $log(1-D(G(z))$ must be minimized. When training the generator $G$, the discriminator $D$ takes a fixed argument, so the objective function does not need to have $D(x)$ terms. The minimization objective function of the generator $G$ in the training process is:

$$\max_D(D,G) = E_{z\sim P_z(z)}[log(1-D(G(z)))] \quad (2)$$

Where, $Ez\sim Pz(z)$ - the fake sample generated from the generator $G$;

By combining the respective objective functions of the discriminator $D$ and the generator $G$, we can get the total objective function of generating the adversarial network:

$$min\max_G V(D,G) = E_{x\sim P_{data(x)}}[logD(x)] \\ + E_{z\sim P_z(z)}[log(1-D(G(z)))] \quad (3)$$

## IV. CGAN BASED ON THE COMBINATION OF CNN AND GAN

### A. Basis of neural network:

The basic unit of this network is the neuron, as shown in Figure 2. Suppose a neuron receives n inputs, the weighted sum of the input signals x obtained by the $z \in R$ neuron is shown in formula (1).

$$z = \sum_{i=1}^{n} w_i x_i + b = w^j x + b \qquad (4)$$

Where w is the weight vector and z is the eigenvector obtained by an activation function.

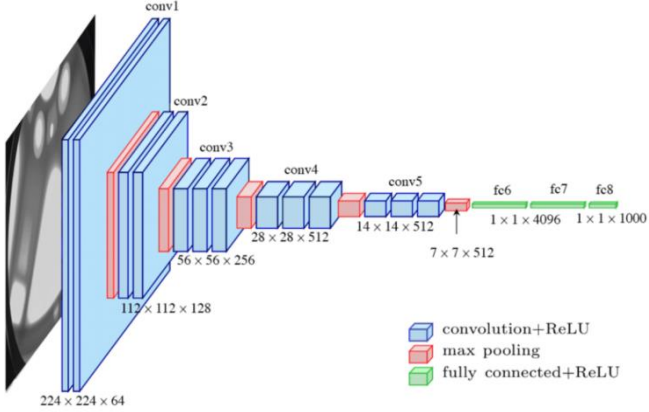

Figure 2 Convolutional neural network

The optimization method of convolutional neural network is a kind of method that can minimize or maximize the objective function. Generally, the derivative order is used in the optimization process according to the optimization algorithm. Gradient descent is the basis for many other optimization algorithms, as shown below:

$$\theta = \theta - \eta \Delta J(\theta) \qquad (5)$$

Where η is the learning rate and $\Delta J(\theta)$ is the gradient of the loss function $J(\theta)$ with respect to the model parameter θ. The loss function $J(\theta)$ is calculated on the whole data set. Therefore, every time the model parameters are updated, the whole data set needs to be calculated, which takes a long time.

### B. CGAN

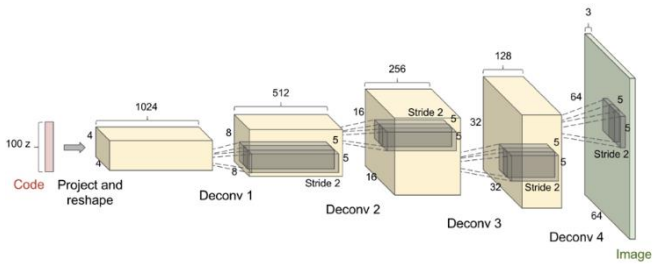

Figure3 CGAN

Several key improvements of CGAN over the original GAN architecture include the complete removal of all pooling layers and the use of transposed convolution in the generation network to implement the up-sampling process to enhance the detail retention of the generated image; In most of the hidden layers of the generator and discriminator (except the output layer of the generator and the input layer of the discriminator), the Batch Normalization technology (BN) is standardized, which effectively accelerates the training process and improves the stability of the model. In the selection of activation function, ReLU function is generally used in each layer of the generator to introduce nonlinearity, while Tanh function is specially used in the output layer to ensure that the output value range is mapped to the interval (-1, 1), which conforms to the characteristics of image pixel value. Instead, each hidden layer of the discriminator adopts the Leaky ReLU function to promote gradient transfer and reduce the problem of gradient disappearance until the last layer applies the Sigmoid function to output a probability value close to the binary classification. These improvements together promote the quality of the generated samples and the convergence performance of the model.

## V. EXPERIMENT

### A. Data set

In this study, we employed the Linked Data[26] methodology to manage a diverse array of publicly available image datasets, ensuring a structured and interoperable data environment crucial for the comprehensiveness and reliability of our research results. Specifically, we utilized classic datasets such as MNIST, which comprises 70,000 grayscale images of handwritten digits 0-9, and CIFAR-10, featuring 6000 32x32 pixel images across 10 color image categories. The application of Linked Data principles allows for the consolidation of varying data formats, enhancing the ability to cross-reference and merge information from these datasets. This is particularly beneficial in fields like machine learning and artificial intelligence, where high-quality data is essential for effectively training models and achieving accurate outcomes. This structured approach not only supports the robustness of our experimental framework but also significantly contributes to the reliability and validity of our findings in image recognition tasks.

### B. Contrast algorithm

Support vector Machine (SVM): SVM is a classification algorithm based on the principle of maximizing the interval, which is not a deep learning method, but performs well in image recognition tasks, especially on small or medium scale data sets[27].

Convolutional Neural Networks (CNNS): CNNS are the most successful and widely used image recognition models in the field of deep learning, efficiently extracting image features through convolutional layers and pooling layers.

Residual network (ResNet): In order to solve the problem of gradient disappearance in deep networks, ResNet introduces residual connections, which makes the network can easily learn hundreds of layers, greatly improving the performance of image recognition.

### C. Experimental result

Table I Experimental results of different models on 4 evaluation indexes

| Model | Precision | Recall | Accuracy | F1 |
|---|---|---|---|---|
| SVM | 0.85 | 0.80 | 0.82 | 0.82 |
| CNN | 0.88 | 0.86 | 0.87 | 0.87 |
| ResNet | 0.92 | 0.91 | 0.91 | 0.92 |
| CGAN | 0.95 | 0.93 | 0.94 | 0.94 |

As can be seen from table I, deep learn-based models (CNN, ResNet, and CGAN) generally outperform traditional SVM methods in image recognition tasks. Among them, SVM performed well in accuracy (0.85), but slightly lower in recall (0.80), overall accuracy (0.82) and F1 score (0.82), indicating limited performance for complex image tasks. CNN significantly improved performance, with accuracy (0.88) and recall (0.86) both higher than SVM, and accuracy (0.87) and F1 score (0.87) showing strong image recognition capabilities. ResNet further improved the accuracy rate (0.92) and recall rate (0.91), and significantly outperformed CNN in overall accuracy (0.91) and F1 score (0.92), indicating superior performance in complex image feature extraction. CGAN, combined with the characteristics of the generated adversarial network, has excellent performance in various indicators, with the highest accuracy rate (0.95), recall rate (0.93), accuracy rate (0.94) and F1 score (0.94), demonstrating its extremely high accuracy and coverage in positive sample recognition and prediction.

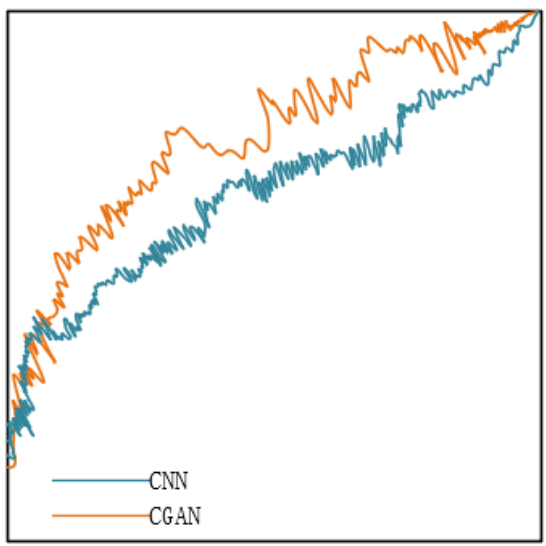

Figure 4 ROC curves of the two models

Figure 4 shows the ROC curves of two different models, CNN and CGAN, which are commonly used to evaluate the performance of classifiers. In this chart, the curves for both "CNN" and "CGAN" extend from the lower left corner to the upper right corner, indicating that both models have some ability to distinguish. Based on the shape of the curve, we can see that the curve of the "CGAN" (orange line) is above the "CNN" (blue line) in most cases. This means that under the same False Positive Rate (FPR), "CGAN" can achieve a higher True Positive Rate (TPR), that is, its overall performance is better than "CNN".

## VI. CONCLUSION

Through comprehensive analysis and experimental verification, this study deeply discusses the image recognition algorithm combining deep learning and generative adversarial network (GAN) and identifies the significant advantages of GAN compared with traditional methods in the field of image recognition. It is confirmed that GAN not only shows superior performance in processing complex image structures but also can capture high-dimensional features and details more effectively, which greatly promotes the development of image recognition technology. In the future, we will further investigate the unique capabilities of GAN in image repair, missing information processing, and high-quality image generation, and broaden its application prospects in image recognition and a wider range of visual tasks. The combination of GAN and deep learning demonstrates the great potential and important value of image recognition technology in the future vision applications of artificial intelligence.